\documentclass[letterpaper, 10 pt, conference]{ieeeconf}  

\IEEEoverridecommandlockouts                              

\usepackage{graphicx}
\usepackage{xcolor}
\usepackage{fancyhdr}
\usepackage{amssymb}
\usepackage{multirow}

\title{\LARGE \bf
Hierarchical Consistency Regularized Mean Teacher for Semi-supervised 3D Left Atrium Segmentation
}

\author{Shumeng Li$^{*1,2}$, Ziyuan Zhao$^{*1,2}$, Kaixin Xu$^{1}$, Zeng Zeng$^{\dagger 1}$, Cuntai Guan$^{2}$
\thanks{* Contributed equally. $^{\dagger}$Corresponding author. This research is supported by Institute for Infocomm Research (I2R), Agency for Science, Technology and Research (A*STAR), Singapore. $^{1}$ Institute for Infocomm Research (I2R), Agency for Science, Technology and Research (A*STAR), Singapore. $^{2}$ Nanyang Technological University, Singapore. This work was done when Shumeng Li was an intern at I2R, A*STAR.}
}

\begin{document}

\maketitle
\thispagestyle{empty}
\pagestyle{empty}

\thispagestyle{fancy}
\fancyfoot{}
\lfoot{\scriptsize{© 2021 IEEE.  Personal use of this material is permitted.  Permission from IEEE must be obtained for all other uses, in any current or future media, including reprinting/republishing this material for advertising or promotional purposes, creating new collective works, for resale or redistribution to servers or lists, or reuse of any copyrighted component of this work in other works.}}

\begin{abstract}

Deep learning has achieved promising segmentation
performance on 3D left atrium MR images. However,
annotations for segmentation tasks are expensive, costly
and difficult to obtain. In this paper, we introduce a
novel hierarchical consistency regularized mean teacher
framework for 3D left atrium segmentation. In each
iteration, the student model is optimized by multi-scale
deep supervision and hierarchical consistency
regularization, concurrently. Extensive experiments have
shown that our method achieves competitive performance
as compared with full annotation, outperforming other
state-of-the-art semi-supervised segmentation methods.
\newline
\indent \textit{Clinical relevance}— The proposed method can efficiently utilize limited labeled data and abundant unlabeled data simultaneously for 3D left atrium segmentation, and achieve comparable performance compared with full annotation, which can be immediately implemented to any medical image segmentation tasks and extended into other clinical applications. 
\end{abstract}

\section{INTRODUCTION}


Deep learning has been widely applied in the field of medical image processing, in which, segmentation techniques play an important role in diagnostic analysis and clinical treatment~\cite{litjens2017survey}. 
Segmenting left atrium (LA) structures from magnetic resonance (MR) images accurately can provide a pre-operative assessment of its anatomy, which is essential for treating various cardiovascular diseases, such as atrial fibrillation~\cite{lang2015recommendations}. 
Despite the promising performance of deep learning on LA segmentation~\cite{ronneberger2015u,milletari2016v}, most of the existing methods rely on a great number of expensive annotations delineated by experienced experts, and the labeling process is tedious and laborious.


To ease the manual labeling burden,  significant effort has been devoted to utilizing available annotations and resources efficiently to improve the segmentation performance~\cite{cheplygina2019not,tajbakhsh2020embracing}. Noting that plenty of unlabeled data can be easily collected, many works focus on leveraging limited data with annotations and abundant data without annotations together for improving the segmentation performance in a semi-supervised learning (SSL) manner~\cite{bai2017semi,asdnet2018,zhao2019semi,tcse2020}.
In recent years, the mean teacher (MT) model~\cite{tarvainen2017mean} has achieved great success in SSL, which enforces the consistency of the predictions with different perturbed inputs between student and teach models thereby improving the robustness of the student model. 
However, the consistency is only applied to the final predictions. Following~\cite{tarvainen2017mean}, many perturbation-based SSL methods have been proposed, obtaining promising performance on LA segmentation~\cite{uamt2019,LG-ER-MT2020,DUWM2020}. These methods focus on uncertainty estimations to exploit rich unlabeled data from different perspectives with high computational complexity. For instance, UA-MT~\cite{uamt2019} estimates the uncertainty with multiple forward passes via Monte Carlo Dropout~\cite{NIPS2017_2650d608}. Another line of research~\cite{li2020shape, DTC2021} tends to build task-level regularization by adding additional auxiliary tasks. In the absence of expert knowledge, these methods easily suffer from negative transfer.



In this work, a novel mean teacher-based SSL framework is presented for 3D LA segmentation, in which, we encourage the prediction consistency at different scales between the student and teacher networks. 
Concretely, we first extend the backbone framework to generate multi-scale predictions. Based on that, we inject multi-level deep supervision to different layers with labeled data to optimize the student network thereby producing more reliable targets. Meanwhile, we enforce the prediction consistency between student and teacher networks hierarchically to further regularize the student network with unlabeled data. Consequently, the proposed framework can efficiently take advantage of both labeled and unlabeled data for accurate SSL LA segmentation. 
Our framework is extensively evaluated on MICCAI 2018 Atrial Segmentation Challenge dataset~\cite{xiong2021global}. 
Experimental results demonstrate that the proposed approach can effectively leverage unlabeled data for LA segmentation, and outperform many state-of-the-art SSL techniques.



\begin{figure*}[htb]
\centering
\includegraphics[width=0.87\textwidth]{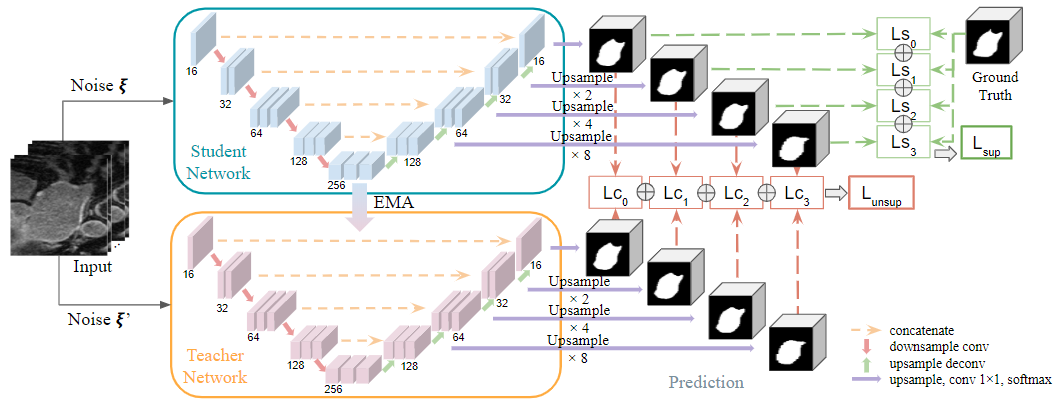}
\caption{Overview of our proposed framework. The student network is trained by minimizing $L_{sup}$ weighted by the hierarchical supervised loss and $L_{unsup}$ weighted by the hierarchical consistency loss. }
\label{fig:pipeline}
\end{figure*}

\section{RELATED WORK}

Semi-supervised learning has achieved promising performance on medical image segmentation under the scarcity of labeled data. 
Bai~\emph{et al.}~\cite{bai2017semi} involve a self-training process for cardiac MR image segmentation, where model updates and pseudo annotations are performed alternatively. 
Meanwhile, self-ensembling methods, particularly the mean teacher (MT) model~\cite{tarvainen2017mean}, have received much attention on semi-supervised image segmentation. Specifically, the mean teacher (MT) structure~\cite{tarvainen2017mean} forces the consistency of the predictions with inputs under different perturbations between the student and teacher models and further boosts the model performance.
Following that, Yu~\emph{et al.}~\cite{uamt2019} utilize uncertainty estimations with Monte Carlo sampling~\cite{NIPS2017_2650d608} to generate uncertainty maps for more reliable predictions.
Beyond the prediction consistency, Li~\emph{et al.}~\cite{tcse2020} construct the transformation consistency into the self-ensembling model (TCSM) to improve the model generalization ability.
Hang~\emph{et al.}~\cite{LG-ER-MT2020} design the local and global structural consistency to extract structural information for semi-supervised LA segmentation.
Li~\emph{et al.}~\cite{li2020shape} impose geometric constraints via an adversarial loss for shape-aware segmentation. 
Considering the consistency of different tasks, Luo~\emph{et al.}~\cite{DTC2021} apply a multi-task framework for segmentation and level set function regression together, enforcing the representation consistency at task level.

Despite the great success of semi-supervised segmentation, these methods suffer from high computational complexity, which increases the running time.
It is observed that the consistency of existing methods is only applied to the final outputs of each voxel, leaving the abundant information from hidden layers unexplored. 
On the other hand, many works~\cite{dou20173d,zhao2021dsal} obtain multi-scale predictions from different layers for boosting the network performance.
Inspired by this, we inject deep supervision into hidden layers for accelerating the convergence of model optimization, thereby improving the prediction accuracy.
Moreover, we expect to leverage predictions at different scales to further regularize the mean teacher model hierarchically, while obtaining more reliable outputs via multi-level deep supervision with minimal extra computational cost.

\section{METHODOLOGY}

Fig.~\ref{fig:pipeline} illustrates the proposed mean teacher-based SSL framework.
Our method follows the basic spirit of the mean teacher and takes 3D medical images under different perturbations as inputs and encourages the consistency of the segmentation predictions.
The student network and the teacher network utilize the same backbone structure.
We introduce additional prediction layers to estimate the quality of the hierarchical hidden representations.
On one hand, deeply supervised learning with labeled data is performed by minimizing the multi-level segmentation loss from the different hidden layers and the final layer.
On the other hand, the multi-level segmentation prediction maps of the teacher network are taken as the learning goal of the student network,  which is regularized by hierarchical consistency to effectively take advantage of unlabeled data.

\subsection{Multi-scale Deeply supervised Network}

In our framework, 3D V-Net~\cite{milletari2016fully} is employed as the backbone.
To extract hierarchical hidden representations, we add auxiliary layers after each block of the decoding stage to form a hierarchical supervised component (HS). Specifically, the auxiliary layers consist of an upsampling layer, a 1 $\times$ 1 $\times$ 1 convolution, and a softmax layer to obtain the multi-scale predictions. 
For the labeled dataset $D_L = \left\{ \left(x_i, y_i \right), i = 1, \ldots, L \right\}$, we consider the multi-scale predictions into model optimization for deeply supervised learning. We calculate the loss between predictions at different scales and ground truths respectively to obtain multi-scale deep supervision. Formally, the hierarchical supervised segmentation loss function is defined as:
\begin{equation}
\mathcal{L}_{sup} = \sum\limits_{s=1}^{S} \alpha_{s}  \frac{\mathcal{L}_{dice} \left(f^s \left(x_i \right), y_i \right) + \mathcal{L}_{ce}\left(f^s \left(x_i \right), y_i \right)}{2},
\end{equation}
where $\mathcal{L}_{ {ce }}$ and $\mathcal{L}_{ {dice }}$ are cross-entropy loss and dice loss, respectively. $f^s(\cdot)$ denotes the segmentation prediction of HS at scale $s$. $\alpha_s$ is the weights of HS component at scale s in the hierarchical network.

\subsection{Hierarchical Consistency Regularized Mean Teacher}
In SSL task, the training dataset consists of the labeled data $D_L = \left\{ \left(x_i, y_i \right), i = 1, \ldots, L \right\}$ and the unlabeled data $D_U = \left\{x_i, i = L+1, \ldots, L+U \right\}$, where $x_i \in \mathbb{R}^{H \times W \times D}$ is the input 3D volume and $y_i \in \left\{0, 1 \right\}^{H \times W \times D}$ is the 3D ground truth. $L$ and $U$ denote the number of 3D volumes of the labeled dataset and unlabeled dataset, respectively.


Following the design of the mean teacher~\cite{uamt2019},
the student network and the teacher network share the same network backbone, and the teacher network weights ${\theta}'$ are updated with the exponential moving average (EMA) of the student network weights $\theta$. 
The teacher network weights update strategy at step $t$ is expressed as ${\theta}'_t = \eta {\theta}'_{t-1} + \left(1 - \eta\right) \theta_t$, where $\eta$ is to control the EMA update rate, weighing the influence of the teacher network by the weights of current student network and the teacher network at step $t-1$.  

For the unsupervised dataset, we expect the segmentation results of the student network and teacher network to be consistent. To further enforce the consistency between the two networks, we design a hierarchical unsupervised consistency component (HU) to encourage the prediction consistent in the hidden feature space. 
In detail, based on the multi-scale feature maps generated from the student and teacher networks, the hierarchical consistency regularization is introduced. 
The regularization effectively utilizes unlabeled data, and the consistency constraint between the student network and the teacher network is determined as the mean squared error (MSE) loss of multi-scale predictions between them, which is denoted as:
\begin{equation}
\mathcal{L}_{unsup} = \sum\limits_{s=1}^{S} \alpha_{s} \cdot  \left \| f^s \left(x_i; {\theta}', {\xi}' \right) - f^s \left(x_i; \theta, \xi \right) \right \|^2,
\end{equation}
where $\left(\theta, \xi\right)$ and $\left({\theta}', {\xi}'\right)$ denote the different weights and input data with different perturbations of student and teacher networks, respectively.

The unsupervised regularization loss can be combined with the deeply supervised loss to optimize the network together. The training objective can be formulated as:
\begin{equation}
\min_\theta \sum\limits_{i=1}^{L} \mathcal{L}_{sup} \left(x_i \right) + \lambda\left(t\right) \sum\limits_{i=1}^{L+U} \mathcal{L}_{unsup} \left(x_i \right),
\end{equation}
where $\mathcal{L}_{sup}$ is the supervised term and $\mathcal{L}_{unsup}$ is the unsupervised consistency regularization term, respectively. $\lambda\left(t\right)$ is a time-dependent Gaussian weighting function that controls the supervised loss and the regularization term, which is defined as $\lambda\left(t\right) = 0.1 \cdot e^{-5{\left(1 - \frac{t}{t_{max}} \right)}^2}$, where $t$ and $t_{max}$ denote the current and maximum training steps, respectively.

\section{EXPERIMENTS}
\subsection{Dataset and Experimental Settings}

We evaluate our proposed semi-supervised method on the dataset of 2018 Atrial Segmentation Challenge for left atrium segmentation in 3D gadolinium-enhanced MR image scans (GE-MRIs)~\cite{xiong2021global}. 
The dataset consists of 100 scans with segmentation masks and the resolution of the dataset is $0.625 \times 0.625 \times 0.625$ $mm^3$.
Following the previous work~\cite{uamt2019, li2020shape, LG-ER-MT2020, DUWM2020, DTC2021}, we select 80 samples for training and 20 for testing and use the same data preprocessing methods for a fair comparison.

The proposed architecture is implemented in Pytorch and updated by the SGD optimizer for $6000$ iterations in total by setting the initial learning rate of $0.01$ and a stepped decay by $0.1$ every $2500$ iterations in our experiments. Hierarchical classifier mechanism is introduced to the V-Net backbone and the number of scales $S$ is $4$. 
The scale-wise weights $\left\{ \alpha_s, s = 0, \ldots, 3 \right\}$ for supervised loss and unsupervised consistency loss are specified as $\left\{ 0.5, 0.4, 0.05, 0.05 \right\}$, empirically. 
Following Tarvainen and Valpola~\cite{tarvainen2017mean}, the decay parameter $\eta$ of exponential moving average (EMA) is $0.99$.
We set the batch size to $4$ and each batch consists of two labeled volumes and two unlabeled volumes. 
For evaluation, four metrics employed to quantitatively evaluate the segmentation performance are Dice, Jaccard, the average surface distance (ASD), and the $95\%$ Hausdorff Distance ($95$HD).

\subsection{Results and Discussions}

\begin{table}[t]
\caption{Segmentation results of different approaches.}
\label{tab:results}
\centering
\begin{tabular}{c|c|c|c|c|c} 
\hline
\multirow{2}{*}{Method} & Scans Used & \multicolumn{2}{c|}{Metrics (\%)~$\uparrow$} & \multicolumn{2}{c}{Metrics (Voxel)~$\downarrow$}  \\ 
\cline{2-6}
                        & L / U      & Dice  & Jaccard                   & ASD  & 95HD                          \\ 
\hline
V-Net                   & 80 / 0     & 91.14 & 83.32                     & 1.52 & 5.75                          \\ 
V-Net                   & 16 / 0     & 86.03 & 76.06                     & 3.51 & 14.26                         \\ 
\hline
ASDNet~\cite{asdnet2018}                  & 16 / 64    & 87.90 & 78.85                     & 2.08 & 9.24                          \\
TCSM~\cite{tcse2020}                    & 16 / 64      & 88.15 & 79.20                     & 2.44 & 9.57                          \\
MT~\cite{tarvainen2017mean}                      & 16 / 64    & 88.12 & 79.03                     & 2.65 & 10.92                         \\
UA-MT~\cite{uamt2019}                   & 16 / 64    & 88.88 & 80.21                     & 2.26 & 7.32                          \\
SASSNet~\cite{li2020shape}                 & 16 / 64    & 89.27 & 80.82                     & 3.13 & 8.83                          \\
DTC~\cite{DTC2021}                     & 16 / 64    & 89.42 & 80.98                     & 2.10 & 7.32                          \\
LG-ER-MT~\cite{LG-ER-MT2020}                & 16 / 64    & 89.62 & 81.31                     & 2.06 & 7.16                          \\
DUWM~\cite{DUWM2020}                    & 16 / 64    & 89.65 & 81.35                     & \textbf{2.03} & 7.04                          \\ 
\hline
\textbf{Ours}           & 16 / 64    & \textbf{90.04} & \textbf{81.98}                     & 2.18 & \textbf{6.93}                           \\
\hline
\end{tabular}
\end{table}

\begin{figure}[thb]
\centering
\includegraphics[width=0.48\textwidth]{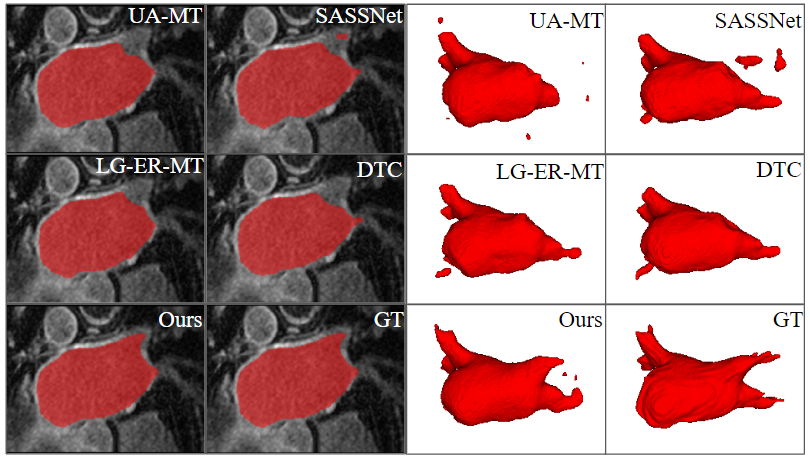}
\caption{Example of 2D and 3D results of five semi-supervised segmentation methods, where GT denotes corresponding ground truth labels. }
\label{fig:example}
\end{figure}

In our experiments, we use $16$ out of $80$ training samples as labeled dataset $D_L$ and the rest 64 samples are served as unlabeled dataset $D_U$.
The results in Table~\ref{tab:results} compare our proposed approach and several other previous image segmentation methods.
It is observed that there are significant improvements in segmentation performance for semi-supervised methods when compared to the supervised-only method (V-Net) for the same amount of labeled data. 
In addition, the results show that our method achieved up to $90.04\%$ in Dice score. 
It is noticed that our approach is superior to other semi-supervised methods and obtains comparable segmentation results using only $1/5$ labeled data as compared with full annotation.

Figure~\ref{fig:example} exemplifies the visualization results of UA-MT~\cite{uamt2019}, SASSNet~\cite{li2020shape}, LG-ER-MT~\cite{LG-ER-MT2020}, DTC~\cite{DTC2021}, our network and ground truth segmentation. 
It is observed that our method can obtain more reliable predictions which are closer to the ground truth segmentation labels.
Remarkably, through our method, a more complete shape of the left atrium and more details are predicted, especially in the boundary area. 
This further demonstrates the effectiveness of our proposed strategy for 3D left atrium image segmentation.

Furthermore, ablation studies are adopted to evaluate the effectiveness of the hierarchical supervised component (HS) and the hierarchical unsupervised consistency component (HU).
We list the results in Table~\ref{tab:ablation}. 
Compared with the V-Net trained on labeled data only, multi-scale deep supervision,~\emph{i.e.,} V-Net + HS can efficiently improve the segmentation performance. 
And it can also be employed for SSL to better utilize the labeled data. 
Then we introduce HU and HS to the mean teacher model, individually, \emph{i.e.,} MT + HU and MT + HS, respectively. The results demonstrate the effectiveness of the two components. Lastly, combining two components together can achieve better performance.

\begin{table}[t]
\caption{Effectiveness of core component in the proposed framework, where HS and HU respectively denote the hierarchical supervised term and unsupervised consistency term.}
\label{tab:ablation}
\centering
\begin{tabular}{c|c|c|c|c|c}
\hline
\multirow{2}{*}{Method} & Scans Used & \multicolumn{2}{c|}{Metrics (\%)~$\uparrow$} & \multicolumn{2}{c}{Metrics (voxel)~$\downarrow$} \\ \cline{2-6} 
                        & L / U      & Dice           & Jaccard          & ASD              & 95HD              \\ \hline
V-Net                   & 16 / 0     & 86.03          & 76.06            & 3.51             & 14.26             \\
V-Net + HS              & 16 / 0     & 86.25          & 76.45            & 2.89             & 11.07             \\ \hline
MT                      & 16 / 64    & 88.12          & 79.03            & 2.65             & 10.92             \\
MT + HU                 & 16 / 64    & 88.16          & 79.09            & 2.95             & 10.62             \\
MT + HS                 & 16 / 64    & 89.32          & 80.87            & 2.70             & 8.73             \\
MT + HU + HS            & 16 / 64    & \textbf{90.04}          & \textbf{81.98}            & \textbf{2.18}             & \textbf{6.93}             \\ \hline
\end{tabular}
\end{table}

\section{CONCLUSION}

In a nutshell, a novel and simple approach is proposed for semi-supervised 3D medical image segmentation. 
We present a framework based on the mean teacher architecture and it encourages the consistency of predictions between the student and the teacher networks with the same input under different perturbations.
We perform the hierarchical prediction consistency based on hidden feature space to regularize the mean teacher model while obtaining more reliable outputs via multi-scale deep supervision. 
Experimental results on the Atrial Segmentation Challenge dataset demonstrate the superiority of our framework over the other state-of-the-art semi-supervised learning methods. 
Besides, the proposed method is simple and versatile, which can be widely applied to other segmentation tasks.

\bibliographystyle{IEEEbib}
\bibliography{refs1.bib}

\end{document}